\documentclass[journal]{IEEEtran}
\usepackage{algorithmicx,algorithm}
\usepackage[scaled=1.0]{helvet}
\usepackage{times}
\usepackage{graphicx}
\usepackage{subfigure}
\usepackage{parskip}
\usepackage{color}
\usepackage{caption2}
\usepackage{multirow}
\usepackage{bm}
\usepackage{amsmath,amsfonts,amssymb,amsthm}
\usepackage{enumerate}
\usepackage{paralist}
\DeclareMathOperator*{\argmin}{arg\,min} % thin space, limits underneath in displays
\newcommand{\etal}{\textit{et al}. }
\newcommand{\ie}{\textit{i}.\textit{e}.}

\begin{document}
	\title{High Speed Tracking With A Fourier Domain Kernelized Correlation Filter}
	
	\author{
		\vskip 1em
		{
			Mingyang Guan,
			Zhengguo Li, \emph{Senior Member, IEEE},
			Renjie He, and
			Changyun Wen, \emph{Fellow, IEEE}
		}
		
		\thanks{
			
			{	
				M. Guan, R. He, and C. Wen are with the School of Electrical and Engineering, College of Engineering, Nanyang Technological University, Singapore 639798 (e-mail: guan0050@e.ntu.edu.sg; renjie.he@ntu.edu.sg; ecywen@ntu.edu.sg;).
				
				Z. Li is with the Department of Signal Processing, Institute for Infocomm Research, Singapore 138632 (e-mail:ezgli@i2r.a-star.edu.sg).
			}
		}
	}

	
	\markboth{}%Journal of \LaTeX\ Class Files}%,~Vol.~6, No.~1, January~2007}%
	{Shell \MakeLowercase{\textit{et al.}}: Bare Demo of IEEEtran.cls
		for Journals}
	
	
	
	\maketitle
	
	\begin{abstract}
		It is challenging to design a high speed tracking approach using $\ell_1$-norm due to its non-differentiability.
		In this paper, a new kernelized correlation filter is introduced by leveraging the sparsity attribute of $\ell_1$-norm based regularization to design a high speed tracker.
		We combine the $\ell_1$-norm and $\ell_2$-norm based regularizations in one Huber-type loss function, and then formulate an optimization problem in the Fourier Domain for fast computation, which enables the tracker to adaptively ignore the noisy features produced from occlusion and illumination variation, while keeping the advantages of $\ell_2$-norm based regression.
		This is achieved due to the attribute of Convolution Theorem that the correlation in spatial domain corresponds to an element-wise product in the Fourier domain,
		resulting in that the $\ell_1$-norm optimization problem could be decomposed into multiple sub-optimization spaces in the Fourier domain.
		But the optimized variables in the Fourier domain are complex, which makes using the $\ell_1$-norm impossible if the real and imaginary parts of the variables cannot be separated. 
		However, our proposed optimization problem is formulated in such a way that their real part and imaginary parts are indeed well separated.
		As such, the proposed optimization problem can be solved efficiently to obtain their optimal values independently with closed-form solutions.
		Extensive experiments on two large benchmark datasets demonstrate that the proposed tracking algorithm significantly improves the tracking accuracy of the original kernelized correlation filter (KCF \cite{1heri2015}) while with little sacrifice on tracking speed.
		Moreover, it outperforms the state-of-the-art approaches in terms of accuracy, efficiency, and robustness.
	\end{abstract}
	
	
	% for peerreview papers, inserts a page break and creates the second title.
	% Will be ignored for other modes.
	\IEEEpeerreviewmaketitle
	\section{Introduction}
	Visual object tracking is one of the fundamental research problems in computer vision, with a wide range of applications such as vehicle tracking and robotic navigation.
	Despite great progress has been made in recent years \cite{1heri2015, 1bolme2010, 1danelljan2013, 1Hare2016, 1Zhang2014, 1Chen2017}, due to numerous factors in real applications such as computational cost, there are still some challenging problems when a practical visual tracking system is designed to meet low computational cost requirements from an embedded system like robotic system for example.
	
	Recently, correlation filter was introduced into visual community, which has pushed forward the research on visual object tracking.
	Many correlation filter based trackers were proposed, see \cite{1heri2015, 1bolme2010, 1danelljan2013} for examples, and achieved satisfactory performance on robustness, accuracy and tracking speed.
	Bolme \etal \cite{1bolme2010} firstly proposed a correlation filter based tracker, called Minimum Output Sum of Squared Error (MOSSE).
	Henriques \etal \cite{1heri2015} improved the performance of MOSSE by taking advantage of the Kernel Trick to classify on richer non-linear feature spaces, that MOSSE could not handle.
	Danelljan \etal \cite{1danelljan2013} further improved the correlation filter based tracker by applying the technology in the MOSSE to handle the scale variation of the object.
	To prevent the training from overfitting, a $\ell_2$-norm regularization term is defined in \cite{1heri2015} and \cite{1danelljan2013}.
	It is well known that $\ell_2$-norm regularization based regression guarantees high computational efficiency because an analytical solution can be obtained.
	However, the target appearance may change significantly in various situations, such as occlusion and illumination variation.
	A robust regularization is required to keep the tracker responding to these challenges reliably, and at the meantime avoid the overfitting.
	
	Compared with $\ell_2$-norm, the $\ell_1$-norm is more robust to resistance the outliers as it is able to ignore extreme coefficients.
	In addition, the $\ell_1$-norm can achieve a sparsity parameter space, which could improve the discriminative of the target against the surrounding background.
	A detail comparison between them is shown in Table \ref{tab:norm}.
	Based on Table \ref{tab:norm}, it might be a good trial to combine the advantages of $\ell_1$-norm with those of $\ell_2$-norm to handle various challenges in visual tracking.
	This is triggered by the fact that lots of pixels near the target are noisy due to occlusion and illumination variation.
	These distractions will produce fake responses over the search region, leading to an inaccurate target location.
	\begin{table}[!h]
		\caption{Differences of $\ell_2$ regression and $\ell_1$ regression}
		\begin{center}
			\tabcolsep 0.06in
			\renewcommand{\arraystretch}{1.5}
			\begin{tabular}{c|c|c}
				\hline
				& $\ell_2$ regression & $\ell_1$ regression \\\hline
				robustness & not very robust & robust\\\hline
				stability & stable solution & unstable solution\\\hline
				number of solution & one & possible multiple\\\hline
				feature selection & no & built-in\\\hline
				sparsity & non-sparse & sparse\\\hline
				complexity & low (analytic solution) & high\\\hline
			\end{tabular}
		\end{center}
		\label{tab:norm}
	\end{table}
	
	In the past years, various research works were conducted to build a robustness visual tracker by exploiting the advantage of $\ell_1$-norm.
	For example in \cite{1mei2009} and \cite{1bao2012}, sparse representation was explored by solving a $\ell_1$-norm regularized least squares problem.
	Sui \etal \cite{1yao2016} studied the influence of $\ell_1$-loss and $\ell_2$-loss functions on the performance of correlation filter based tracker.
	They proposed three different sparsity related loss functions and an iterative method is adopted to approximate the $\ell_1$-norm based optimization problem.
	Similar to \cite{1yao2016}, we also aim to improve the robustness of the correlation filter based tracker by taking the advantage of the sparsity attribute of $\ell_1$-norm.
	Due to the non-differentiability of $\ell_1$-norm, it is usually difficult to obtain a closed-form solution for the optimization problem.
	
	Inspired by the Huber loss function \cite{1huber1964} which combines the squared loss with absolute-value loss function, a novel regularization term with sparsity constraint is introduced in this paper to enhance the robustness and accuracy of the visual tracker.
	The optimization problem is formulated in the Fourier Domain for fast computation.
	This can be achieved due to the attribute of Convolution Theorem that the correlation in spatial domain corresponds to an element-wise product in the Fourier domain.
	Thus the $\ell_1$-norm optimization problem could be decomposed into multiple sub-optimization spaces in the Fourier domain and be solved efficiently to avoid the time-consuming process.
	On the other hand, the optimized variables in the Fourier domain are complex.
	It is impossible to use the $\ell_1$-norm if the real and imaginary parts of the variables cannot be separated.
	Fortunately, they are well separated in the proposed optimization problem formulation, which enables their optimal values to be obtained independently.
	Due to the convex and differentiable characteristics of the Huber loss function,
	an efficient closed-form solution is given and discussed for the proposed regularization terms in this paper.
	The proposed tracker keeps advantages of both $\ell_1$- and $\ell_2$- norms.
	The scale estimation in the DSST \cite{1danelljan2013} was adopted to further improve the performance of the proposed tracker.
	
	With the proposed novel solution for sparsity constraint regularization term, the tracker is able to obtain a more robust tracking performance compared with most popular correlation filter based trackers developed recently.
	Our proposed scheme is tested on frequently used benchmark datasets OTB-50 \cite{wu2013online}.
	It is shown from the experimental results presented in Section \ref{sec:performance} that the proposed tracking method yields significant improvements over the state-of-the-art trackers under various evaluations conditions, while the tracking speed for the proposed tracker is reduced little compared with KCF \cite{1heri2015} and DSST \cite{1danelljan2013}.
	Therefore, the proposed tracker can be adopted to design a real time navigation system for mobile robots without any precise maps.
	The visual tracker will be used to determine the moving direction for a robot, just like head-orientation cells for mammals \cite{geva2015spatial}.
	The rest of this paper is organized as below.
	Relevant works are presented in Section \ref{sec:relate}.
	Its performance is evaluated in Section \ref{sec:performance}.
	Finally, concluded remarks are given in Section \ref{sec:conclusion}.
	
	\section{Related Works}
	\label{sec:relate}
	In this section, we briefly review tracking methods closely related to this work.
	\subsection{Tracking-by-detection}
	The tracking problem in some cases are treated as a tracking-by-detection problem, which repeatedly detects the target in a local/global window.
	A binary classification is learned online to find a decision boundary which has highest similarity with the given target.
	Lots of attentions were paid to learn a more discriminative model with less ambiguity to reduce model drifts, thus improve the tracking accuracy, such as, multiple instance learning \cite{1Babenko2011}, semi-supervised learning \cite{1Grabner2008}, support vector machine (SVM) \cite{1Hare2016, 1Zhang2014, 1Chen2017} and P-N learning \cite{1Kalal2012} for examples.
	Avidan \cite{1avidan2007} combined the SVMs and boosting algorithms as classifiers respectively for visual tracking.
	Babenko \etal \cite{1Babenko2011} proposed to address the ambiguity of target appearance by applying an online instance learning algorithm to train a discriminative model from multiple instances of the target.
	Kalal \etal \cite{1Kalal2012} decomposed the tracking task into tracking, learning, and detection, where the detection model is online trained from a random forest method.
	Hare \etal \cite{1Hare2016} employed a kernelised structured output SVM to explicitly express the tracking problem as a joint structured output prediction.
	Zhang \etal \cite{1Zhang2014} proposed to learn a multi-expert restoration scheme where each expert is constituted with its historical snapshot, thus the best expert is selected to locate the position of target based on a minimum entropy criterion.
	However, the tracking efficiency is a possible issue for the trackers based on tracking-by-detection strategy, due to plenty of candidate samples needed to be classified at each tracking frame while the tracking speed is significantly important for some applications, such as real time robotic navigation without a precise map.
	
	\subsection{Correlation Filter (CF) Based Tracking}
	Recently, correlation filters were widely used in visual tracking \cite{1bolme2010, 1Bertinetto2016, 1heri2015, 1danelljan2013, 1Ma2015, 1Danelljan2015,2Danelljan2015,4Danelljan2016, 1Mueller2017,1Guan2018}.
	CF-based trackers regress all the circular-shifted samples of input features to a Gausssian shaped regression labels with high computational efficiency
	based on the property of the circulant matrix in the Fourier domain so as to transfer the time-consuming convolutional operation to element-wise product, which at the meantime achieves satisfactory tracking performance in computational efficiency and accuracy.
	Bolme \etal \cite{1bolme2010} initially applied correlation filter to visual tracking by minimizing the total squared error on a set of gray-scale patches.
	Henriques \etal \cite{1heri2013} improved the performance by exploiting the circulant structure of adjacent image patches to train the correlation filter. Further improvement was achieved by proposing kernelized correlation filters (KCF) using kernel-based training and histogram of oriented gradients (HOG) features \cite{1heri2015}.
	Danellijan \etal \cite{1danelljan2013} proposed adaptive multi-scale correlation filters to cover the scale change of the target.
	The trackers in \cite{1heri2015} and \cite{1danelljan2013} can be regarded as the baseline of CF-based trackers, which push forward the research on visual object tracking.
	Many research works were conducted to improve the performance of CF trackers by applying additional discriminate model \cite{1Ma2015, 1Hong2015, 1Bertinetto2016, 1Guan2018}, incorporating with convolutional neural network \cite{2Ma2015, 2Danelljan2015, 3Danelljan2017} and reformulating the optimal function with new characteristics \cite{1Tang2015, 1Liu2016, 1Mueller2017, 1Sui2018}.
	
	\paragraph{CF trackers with additional discriminate model}
	Some researchers improved performance of CF-based tracker by introducing additional discriminate models to alleviate the model drifts \cite{1Ma2015, 1Hong2015, 1Bertinetto2016, 1Guan2018}, through averting poor model updating or re-locating the target and reinitializing tracking model.
	In this case, the tracker can tolerate certain amount of consecutive noisy sequences, including occlusion, motion blur and so on.
	Ma \etal \cite{1Ma2015} estimated the confidence of current tracking state based on the response from correlation filter to detect tracking failures and a random forest classifier is trained to re-detect the target.
	Bertinetto \cite{1Bertinetto2016} complemented the correlation filter with a colour statistics model of the target to address the deformation sensitive of CF-based tracker.
	Hong \etal \cite{1Hong2015} introduced a biologic-based model to maintain a correlation filter-based short-tracker and a long-term memory of SIFT key-points for detecting the target.
	Guan \etal \cite{1Guan2018} built up an occlusion and tracking failure detection model to identify the tracking failure cases of CF-based tracker, with an event-triggered mechanism to decide if or not to re-locate the target.
	However, these kind of methods improve the CF-based tracker with the support from additional discriminate model, while the original model of correlation filter is usually kept unchanged.
	
	\paragraph{CF trackers with convolutional neural network}
	On the other hand, some approaches improved the CF-based tracker by leveraging some stronger features extracted from neural network for a richer representation of the tracking target \cite{2Ma2015, 2Danelljan2015, 3Danelljan2017, 4Danelljan2016}.
	Ma \etal \cite{2Ma2015} employed multiple convolutional layers in a hierarchical ensemble of independent discriminative correlation filter (DCF) based trackers.
	Danelljan \etal \cite{2Danelljan2015} used the output of first convolutional layer of a CNN as the features to represent the target and apply it in a discriminative correlation filter based tracking framework.
	The performance of convolution operation based DCF trackers were further improved in \cite{3Danelljan2017, 4Danelljan2016}.
	However, convolution neural network based tracker is hardware dependent, \ie, the supporting from GPU, and thus not suit for the current applications on robotics due to the highly power cost.
	
	\paragraph{CF trackers with reformulating the optimal function}
	New approaches were proposed to improve the CF trackers by reformulating the optimization function by taking into consideration with new characteristics \cite{1Tang2015, 1Danelljan2015, 1Liu2016, 1Mueller2017, 1Sui2018}.
	Tang \etal \cite{1Tang2015} derived a multi-kernel correlation filter which takes the advantage of the invariance-discriminative power spectrum of various features.
	Danellijan \etal \cite{1Danelljan2015} applied a spatial weight on the regularization term to address the boundary effects, thus greatly enlarged the searching region and improved the tracking performance.
	Liu \etal \cite{1Liu2016} reformulated the CF tracker as a multiple sub-parts based tracking problem, and exploited circular shifts of all parts for their motion modeling to preserve target structure.
	Mueller \etal \cite{1Mueller2017} reformulated the original optimization problem by incorporating the global context within CF trackers.
	Sui \etal \cite{1Sui2018} proposed to enhance the robustness of the CF tracker by adding a $\ell_1$ norm regularization item with the original optimization problem, and an approximate solution for $\ell_1$ norm was given.
	Similar to the idea in \cite{1Sui2018} that formulated a sparsity constraint to alleviate the influence from distractive features produced by occlusion and deformation for examples, we also propose to introduce the $\ell_1$ norm sparsity constraint for the CF tracker.
	However, different from \cite{1Sui2018} that formulates the optimization problem in spatial domain and layouts an approximation solution, we reformulate the CF optimization problem in the the Fourier domain to efficiently solve the $\ell_1$ norm optimization problem, with the help of Convolution Theorem.
	
	\section{The proposed KCF in the Fourier Domain}
	\label{sec:propose}
	In this section, a new kernelized correlation filter is proposed in the Fourier domain by combining advantages of both $\ell_2$- norm and $\ell_1$- norm based regressions.
	
	\subsection{Fourier Domain kernelized Correlation Filter }
	Define a non-linear feature mapping function $\psi: R^n \mapsto R^d (d\gg n)$, the
	kernel trick is to find the inner product of feature mapping
	without calculating the high dimension features explicitly.
	The kernel function is defined as $\kappa: R^n\times R^n \mapsto R$, such that
	$\kappa(\mathbf{x}_j, \mathbf{x}_{j'}) = \psi^T(\mathbf{x}_j)\psi(\mathbf{x}_{j'})$. Given a test image $\mathbf{z}\in  R^n$ and its
	desired correlation output $\mathbf{g}\in R^m$, the kernel correlation is
	defined in the Fourier domain as:
	\begin{equation}
	\label{contraint}
	\hat{\mathbf{g}} = \hat{\kappa}_Z(\mathbf{x})\odot \hat{\mathbf{h}}^*,
	\end{equation}
	where $\hat{\mathbf{g}}=F(\mathbf{g})$ denotes the DFT of the generating vector $\mathbf{g}$. $\hat{\mathbf{h}}^*$ is the complex-conjugate of $\hat{\mathbf{h}}$. $\hat{\kappa}_Z(\mathbf x)=[\hat{\kappa}(\mathbf x, \mathbf{z}_0), \cdots, \hat{\kappa}(\mathbf x, \mathbf{z}_{n-1}]^T$, and $\mathbf{z}_i(0\leq i\leq n-1)\in R^n$ are the sample-based vectors to imply the connotation of training objective and generated from the training sample $\mathbf z$.
	
	The correlation output is transformed back into the spatial domain using the inverse FFT. When a set of training samples $\mathbf{z}^j$ and their associated
	training outputs $\mathbf{g}_j$ are given, a filter $\mathbf h$ is required to satisfy Eq. (\ref{contraint}). Training is conducted in the Fourier domain to take advantage of the simple element-wise
	operation between the input and the output. To find the filter $\mathbf h$ that maps training inputs to the desired outputs, the sum of squared error (SSE) between the correlation output
	and the desired output in Fourier domain is minimized. One term of the cost function is then defined as
	\begin{equation}
	\label{datafidelityterm}
	E_1 = \frac{1}{2}\sum_j\|\hat{\kappa}_{Z^j}(\mathbf{z}^j)\odot\hat{\mathbf h}^*-\hat{\mathbf g}^j\|_2^2,
	\end{equation}
	and this term is usually called a data fidelity term.
	
	Due to the element-wise product, each item $\hat{\text{h}}_l^*$ in $\hat{\mathbf h}^*$ is independent of the others which allows the optimization to be conducted separately.
	For simplicity, the $l$th elements of $\hat{\kappa}_{Z^j}(\mathbf{z}^j)$, $\hat{\mathbf g}^j$ and $\hat{\mathbf h}^*$ are respectively denoted as
	\begin{eqnarray}
	\label{eq:sub}
	&&\hspace{-7mm}\hat{\kappa}_{Z^j_l}(\text{z}^j_l)=a_l^j+b_l^j*i, \\
	&&\hspace{-7mm} \hat{\text{g}}_l^j = c_l^j+d_l^j*i,\\
	&&\hspace{-7mm}\hat{\text{h}}^*_l = e_l+f_l*i,
	\end{eqnarray}
	where $i^2=-1$. It can be computed that
	\begin{eqnarray}
	\nonumber
	&&\hspace{-11mm}\frac{1}{2}\|\hat{\kappa}_{Z^j_l}(\text{z}^j_l)\cdot\hat{\text h}^*_l-\hat{\text g}^{j}_l\|^2=\frac{(a_l^j)^2+(b_l^j)^2}{2}e_l^2-(a_l^jc_l^j+b_l^jd_l^j)e_l\\
	&&\hspace{-7mm}+\frac{(a_l^j)^2+(b_l^j)^2}{2}f_l^2-(a_l^jd_l^j-b_l^jc_l^j)f_l+\frac{(c_l^j)^2+(d_l^j)^2}{2}.
	\end{eqnarray}
	
	Clearly, $e_l$ and $f_l$ are separated well in the above equation. Subsequently, it can be derived that
	\begin{eqnarray}
	\nonumber
	&&\hspace{-13mm}E_1 = \sum_{l=0}^{m-1}[\frac{\gamma_{1,l}}{2}e_l^2-\gamma_{2,l}e_l]+\sum_{l=0}^{m-1}[\frac{\gamma_{1,l}}{2}f_l^2-\gamma_{3,l}f_l]\\
	&&\hspace{-5mm}+\sum_{l=0}^{m-1}\sum_j\frac{(c_l^j)^2+(d_l^j)^2}{2},
	\end{eqnarray}
	where $\gamma_{1,l}$, $\gamma_{2,l}$ and $\gamma_{3,l}$ are expressed as follows:
	\begin{eqnarray}
	&&\hspace{-11mm}\gamma_{1,l}=\sum_j[(a_l^j)^2+(b_l^j)^2],\\
	&&\hspace{-11mm}\gamma_{2,l}=\sum_j(a_l^jc_l^j+b_l^jd_l^j),\\
	&&\hspace{-11mm}\gamma_{3,l} = \sum_j(a_l^jd_l^j-b_l^jc_l^j).
	\end{eqnarray}
	
	A regularization term is usually utilized to avoid overfitting.
	Regularization terms in \cite{1heri2015} and \cite{1wang2018} are modeled by using the $\ell_2$-norm.
	For example, the regularization term in \cite{1wang2018} is given as
	\begin{equation}
	E_2=\sum_{l=0}^{m-1}(e_l^2+f_l^2).
	\end{equation}
	
	In this paper, a new regularization term is introduced on the basis of differences between the $\ell_2$- and the $\ell_1$- norm based regressions in Table \ref{tab:norm}. 
	Both the advantages of the $\ell_2$ norm based regression and the $\ell_1$ norm based regression are utilized in the proposed term which is defined as follows:
	\begin{equation}
	\label{regularizationterm}
	E_2 = \sum_{l=0}^{m-1}(\phi(e_l)+\phi(f_l)),
	\end{equation}
	where the function $\phi(u)$ is defined as
	\begin{equation}
	\label{phifunction}
	\phi(u)=\left\{\begin{array}{ll}
	|u|; &\mbox{if~}|u|>c\\
	\frac{u^2+c^2}{2c}; &\mbox{otherwise}\\
	\end{array}
	\right.,
	\end{equation}
	and $c$ is a small positive constant.
	
	Inspired by the Huber loss function \cite{1huber1964}, $\phi(u)$ is constructed based on the $\ell_2$ norm if the value of $|u|$ is small and the $\ell_1$ norm  otherwise. 
	It can be easily shown that the function $\phi(u)$ is differentiable and its derivative $\phi'(u)$ is given as
	\begin{equation}
	\phi'(u)=\left\{\begin{array}{ll}
	1; &\mbox{if~}u>c\\
	-1; & \mbox{if~}u<-c\\
	\frac{u}{c}; & \mbox{otherwise}\\
	\end{array}
	\right..
	\end{equation}
	
	Clearly, the function $\phi'(u)$ is a continuous function of $u$. 
	An analytic solution of the proposed filter is thus guaranteed. 
	Moreover, the solution is continuous.
	
	The overall cost function is defined as
	\begin{equation}
	\label{optequation}
	E= E_1+\lambda E_2,
	\end{equation}
	where $\lambda$ is a positive constant.  
	Its role is to obtain a good trade-off between the data fidelity term and the regularization term.
	
	The filter $\hat{\mathbf{h}}^*$ is obtained by solving the following optimization problem:
	\begin{equation}
	\label{optimizationproblem}
	\min_{\hat{\mathbf{h}}^*}\left\{E \right\}.
	\end{equation}
	
	By taking the derivation for (\ref{optimizationproblem}) in terms of $e_l$ and $f_l$ respectively, it can be obtained that
%	\begin{eqnarray}
%	&&\hspace{-11mm}\frac{\partial E}{\partial e_l}=0,\\
%	&&\hspace{-11mm}\frac{\partial E}{\partial f_l}=0,
%	\end{eqnarray}
	\begin{eqnarray}
	&&\hspace{-11mm}\gamma_{1,l}e_l-\gamma_{2,l}+\lambda \phi'(e_l)=0,\\
	&&\hspace{-11mm}\gamma_{1,l}f_l-\gamma_{3,l}+\lambda \phi'(f_l)=0.
	\end{eqnarray}
	
	The optimal values of $e_l$ and $f_l$ are then computed as
	\begin{eqnarray}
	&&\hspace{-3mm}e_l=\left\{\begin{array}{ll}
	\frac{\gamma_{2,l}-\lambda}{\gamma_{1,l}}; &\mbox{if~} \frac{\gamma_{2,l}-\lambda}{\gamma_{1,l}}>c\\
	\frac{\gamma_{2,l}+\lambda}{\gamma_{1,l}}; &\mbox{if~} \frac{\gamma_{2,l}+\lambda}{\gamma_{1,l}}<-c\\
	\frac{c\gamma_{2,l}}{c\gamma_{1,l}+\lambda}; &\mbox{otherwise}\label{eq:el}\\
	\end{array}
	\right.,\\
	&&\hspace{-3mm}f_l=\left\{\begin{array}{ll}
	\frac{\gamma_{3,l}-\lambda}{\gamma_{1,l}}; &\mbox{if~} \frac{\gamma_{3,l}-\lambda}{\gamma_{1,l}}>c\\
	\frac{\gamma_{3,l}+\lambda}{\gamma_{1,l}}; &\mbox{if~} \frac{\gamma_{3,l}+\lambda}{\gamma_{1,l}}<-c\\
	\frac{c\gamma_{3,l}}{c\gamma_{1,l}+\lambda}; &\mbox{otherwise} \label{eq:fl}\\
	\end{array}
	\right..
	\end{eqnarray}
	
	The same as the filters in \cite{1heri2015} and \cite{1wang2018}, an analytic solution is available for the proposed filter. 
	Both the storage and computation can be reduced.
	
	Consider the case that the kernel function $\kappa(\mathbf x,\mathbf{z}_j)$ is defined as the following radial-basis kernel \cite{1heri2015, 1wang2018}:
	\begin{equation}
	\label{kernalnorm}
	\kappa(\mathbf{x},\mathbf{z}_j)=h(\|\mathbf{x}-\mathbf{z}_j\|^2),
	\end{equation}
	where $\mathbf{z}_j = \mathbf{P}^j\mathbf{z}$. $\mathbf{P}$ is a cyclic shift operator which is given in \cite{1davis1994}
%	\begin{equation}
%	\mathbf{P}=\left[\begin{array}{lllll}
%	0 & 0 & 0 & \cdots & 1\\
%	1 & 0 & 0 & \cdots & 0\\
%	0 & 1 & 0 & \cdots & 0\\
%	\vdots & \vdots & \ddots & \ddots & \vdots\\
%	0 & 0 & \cdots & 1 & 0\\
%	\end{array}
%	\right].
%	\end{equation}
	
	To compute the kernel vector efficiently, the norm in Eq. (\ref{kernalnorm}) is expanded as
	\begin{equation}
	h(\|\mathbf{x}-\mathbf{P}^j\mathbf{z}\|^2)=h(\|\mathbf{x}\|^2+\|\mathbf{z}\|^2-2\mathbf{x}\mathbf{P}^j\mathbf{z}).
	\end{equation}
	Since $(\|\mathbf{x}\|^2+\|z\|^2)$ is a constant, the kernel vector can be
	calculated as:
	\begin{equation}
	\kappa_Z(\mathbf{x})=h(\|\mathbf{x}\|^2+\|\mathbf{z}\|^2-2[\mathbf{x}^T\mathbf{z}_1,\cdots,\mathbf{x}^T\mathbf{z}_n]^T).
	\end{equation}
	Using the correlation theory, $\mathbf{x}*Z= [\mathbf{x}^T\mathbf{z}_1,\cdots,\mathbf{x}^T\mathbf{z}_n]^T$. It follows that
	\begin{equation}
	\kappa_Z(\mathbf{x})=h(\|\mathbf{x}\|^2+\|\mathbf{z}\|^2-2F^{-1}(\hat{\mathbf{x}}\odot \hat{\mathbf{z}})),
	\end{equation}
	where $F^{-1}$ denotes the Inverse DFT.
	
	The bottle-neck of the above equation is the forward and backward FFTs, so that the kernel vector can be calculated in complexity $O(n\log n)$.
	
	\subsection{Correlation Filter Learning}
	Object appearance changes among the sequences due to the environment variation such as illumination, motion blur and deformation.
	Hence it is necessary to update the CF filter over time to exploit the temporal information, and avoid the correlation filter changing abruptly in successive frames.
	The learned target appearance $\hat{\mathbf{z}}_t$ in the Fourier Domain and the transformed classifier coefficients $\hat{\mathbf{h}}_t^*$ at the $t$th frame are updated in an incremental manner respectively
	\begin{align}
	&\hat{\mathbf{z}}_t = (1 - \varepsilon)\hat{\mathbf{z}}_{t-1} + \varepsilon\hat{\mathbf{z}}, \nonumber\\
	&\hat{\mathbf{h}}^*_t = (1 - \varepsilon)\hat{\mathbf{h}}^*_{t-1} + \varepsilon\hat{\mathbf{h}}^*,
	\label{eq:update}
	\end{align}
	where $\hat{\mathbf{z}}$ is the DFT of the spatially expanded region according to the position of predicted target at the $(t-1)$th frame,
	$\hat{\mathbf{h}}^*$ is the estimation from (\ref{optimizationproblem}), and $\varepsilon \in (0, 1)$ is a pre-defined model learning rate.
	However, the update for the CF at some frames are unnecessary such as heavy occlusion occurred, since noise samples will cause model pollution.
	One solution for these noise sample is to ignore them based on a predefined metric.
	In this paper, the peak-to-sidelobe ratio (PSR) is adopted to quantify the reliability of the tracked samples.
	Following \cite{bolme2010visual}, the PSR is defined as $P_t = (R_{max} - \mu_t)/\sigma_t$ where $R_{max}$ is the peak values of confidence, $\mu_t$ and $\sigma_t$ are the mean and standard deviation of the response, respectively.
	Such that, the update will only happen when $P_t$ is bigger than a predefined empirical threshold $\varrho_{0}$.
	
	\subsection{Target State Estimation}
	The state estimation for the target includes transformation prediction and scale estimation.
	
	1) Position Estimation:
	At each frame, a search region will be decided based on the previous target state estimation, thus extracts a base sample $\mathbf{z}$.
	A number of candidate samples will be produced with the full circle shifts of $\mathbf{z}$.
	Given the learned model $\mathbf{z}_t$ and $\mathbf{h}_t$, the new position of the target $\mathbf{p}_t$ will be estimated by searching the location with the maximal response of the regression values $f_t(\mathbf{z})$
	\begin{align}
	f_t(\mathbf{z}) = \mathcal{F}^{-1}\left(\hat{\mathbf{h}}^*_t\odot\mathcal{F}(\kappa(\mathbf{z}_t, \mathbf{z}))\right),
	\end{align}
	where $\mathcal{F}$ and $\mathcal{F}^{-1}$ denote the Fourier transform and its inverse,
	$\kappa(\mathbf{z}_t, \mathbf{z})$ is the kernel correlation operator.
	
	2) Scale Estimation:
	To handle the scale problem, an one-dimensional correlation filter is trained on $ N $ image patches with different scales, where pyramid around the estimated position is cropped from the image.
	Assume that the target size in the current frame is $ W\times H $.
	Denote the scale vector as $S=\{\xi^r|r=\lfloor-\frac{N-1}{2}\rfloor, \lfloor-\frac{N-3}{2}\rfloor, ...,\lfloor\frac{N-3}{2}\rfloor,\lfloor\frac{N-1}{2}\rfloor\}$ where $ \xi $ is a parameter representing the base value of scale changing.
	Then the image patch $I_s$ with size $sW \times sH, s\in S$ centered around the previous estimated location $\bm{p}_t$ is resized to a pre-defined template size.
	The HOG feature is used to represent each image patch $I_s$.
	The optimal scale $ \bar{s} $ is given as the highest response from correlation filter
	\begin{align}
	\bar{s} = \argmin_{s \in S} f_s(I_s),
	\label{eq:scale}
	\end{align}
	where $f_s(\cdot)$ denotes the regression values respect to the scale $s$.
	The scale estimation in this paper was used in the DSST \cite{1danelljan2013}.
	The formal description of the proposed tracking method is presented in Algorithm \ref{al:fdkcf}.
	
	\begin{algorithm}
		\caption{The Proposed Fourier Domain Kernelized Correlation Filter}
		\renewcommand\arraystretch{1.15}
		\begin{center}
			\scalebox{1}{
				\begin{tabular}{l}
					\textbf{Input:} Initial target $ B_1 $, Regression objective $\mathbf{g}$,\\
					\hspace{0.4in} and image sequences $ \{I_t\}_1^T $.\\
					\textbf{Output:} estimated bounding states $ \{B_t\}_2^T $.\\
					Initial correlation filter $\hat{\mathbf{z}}_1$ and $\hat{\mathbf{h}}_1$ with initial target $B_1$. \\
					\textbf{for} $ t=2:T $ \textbf{do}\\
					\hspace{0.1in} Compute FFT for the kernel matrix $\hat{\mathbf{K}}=\mathcal{F}(\kappa(\mathbf{z}, \mathbf{z}_{t-1}))$. \\
					\hspace{0.1in} Regression response $f_t(\mathbf{z}) = \mathcal{F}^{-1}\left(\hat{\mathbf{h}}^*_{t-1}\odot\hat{\mathbf{K}}\right)$. \\
					\hspace{0.1in} Locate target position with maximum $ f_t(\mathbf{z})$. \\
					\hspace{0.1in} Estimate target scale from Eq. (\ref{eq:scale}). \\
					\hspace{0.1in} Compute $e_l$ and $f_l$ from Eq. (\ref{eq:el}), Eq. (\ref{eq:fl}), respectively. \\
					\hspace{0.1in} Construct filter $\hat{\mathbf{h}}^*$ with $e_l, f_l, l\in[0, m-1]$. \\
%					\hspace{0.1in} Reform filter $\hat{\mathbf{h}}^* = [e_1+f_1*i, ..., e_m+f_m*i]^T$. \\
					\hspace{0.1in} Update $\hat{\mathbf{z}}_t$ and $\hat{\mathbf{h}}_t$ from Eq. (\ref{eq:update}) with $\hat{\mathbf{h}}^*$ and $\hat{\mathbf{z}}$. \\
					\textbf{end for}\\
				\end{tabular}
			}
		\end{center}
		\label{al:fdkcf}
	\end{algorithm}
	
	\section{Experimental Results}
	\label{sec:performance}
	\subsection{Implementation Details}
	\label{sec:imp}
	The proposed tracker is implemented in MATLAB 2014a and experiments are performed on a PC with $\rm Intel^{\textregistered}$ Xeon(R) E5-1630 3.70GHz CPU and 16GB RAM.
	All parameters of the proposed tracker are kept consistent across all experimental comparisons.
	The training samples are obtained by full circle shifts on the base image patch centered at the current target location.
	HOG features with 31 bins and 4x4 cell size are used to extract the feature for the training samples.
	The Gaussian kernel for HOG feature is set to $\sigma_{\rm HOG} = 0.5$.
	It is worth noting that the Gaussian kernel can be replaced by an edge preserving smoothing filter, such as weighted guided image filter \cite{2li2014}.
	The regularization parameter in (\ref{optequation}) is set to $\lambda = 10^{-5}$.
	We empirically set the parameter $c$ in (\ref{phifunction}) to $50$, the threshold for CF updating $\varrho_{0}$ to 10.
	The scale pool is built based on the suggestion in \cite{1danelljan2013} to contain $33$ different scaling coefficients.
	The datasets used for evaluation are the object tracking benchmark OTB-50 \cite{wu2013online} and VOT2016 \cite{Kristan2016a}, which are two popular datasets used to evaluate the overall performance of the trackers.
	Both datasets contain various challenging sequences for visual tracking, such as occlusion, nonrigid deformation, illumination variation, in-plane rotation, and scale variation.
	
	\subsection{Evaluation on OTB-50 \cite{wu2013online} Dataset}
	The OTB-50 \cite{wu2013online} contains 50 videos involving different objects in their video sequences.
	There are usually two criteria used to evaluate the visual trackers:
	1) distance precision (DP) and 2) overlap precision (OP).
	Regarding the first, a tracking location error $\iota$ indicates the precision of each tracking frame, which is usually defined as the Euclidean distance between the center of the predicted and the ground truth bounding box.
	Then, the DP for a given sequence is defined as the percentage of frames that the tracking location errors are less than the pre-defined threshold $\pi$.
	For success rate, an overlap rate $o$ is usually defined to represent the tracking performance for each tracking frame, which indicates the rate between the overlapped areas and the union areas of the predicted and the ground truth bounding boxes and can be expressed as $\dfrac{A_t\cap A_g}{A_t\cup A_g}$, where $A_t$ and $A_g$ denote the bounding box of the predicted and the labeled.
	Thus, the OP is defined as the percentage of frames that $o$ is greater than a pre-defined threshold $\tau$.
	
	\subsubsection{Comparison with Baseline trackers}
	\label{com:baseline}
\begin{figure}[!htb]
		\centering
		\includegraphics[width=1.45\linewidth]{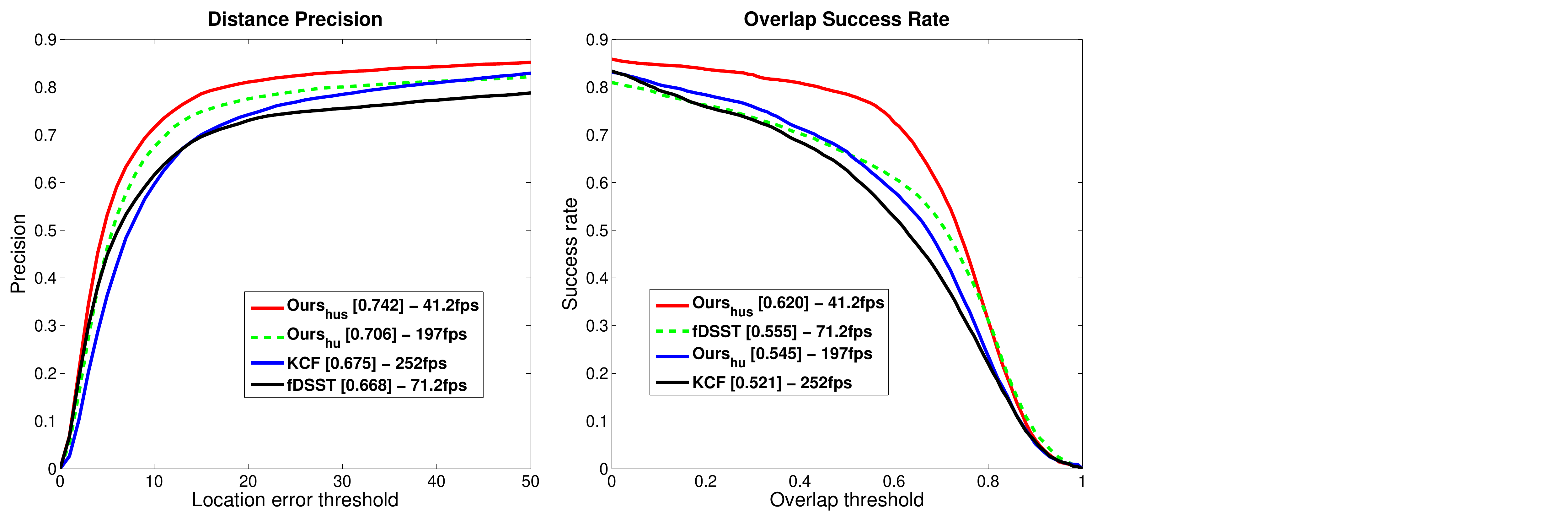}
		\caption{Tracking performance comparison between the proposed tracker and baseline trackers on OTB-50 \cite{wu2013online} benchmark. The scores in the legends indicate the average Area-Under-Curve (AUC) value on all thresholds for distance precision plots and overlap precision plots, respectively.}
		\label{fig:baseline}
	\end{figure}

	To validate the effectiveness of the proposed solution for the Fourier Domain correlation filter with Huber loss function, a comparison experiment with two baseline trackers, named KCF \cite{1heri2015} and fDSST \cite{danelljan2017discriminative}, is designed.
	Note that the proposed Huber loss function directly depends on KCF \cite{1heri2015}, while fDSST \cite{danelljan2017discriminative} is an extension of KCF \cite{1heri2015} by detecting the scale variation of the object.
	Beside, to evaluate the influence of the proposed Huber loss function sophisticatedly, the evaluation for the proposed tracker consists of two parts: 1) implementing the proposed Huber loss function on kernelized correlation filter without predicting the scale variation. We denote this tracker as Ours$_{hu}$.
	2) implementing the proposed Huber loss function at the same time using the method in DSST to predict the scale variation of the object, denoted as Ours$_{hus}$.
	The comparison results are shown in Figure \ref{fig:baseline}.
	
	Figure \ref{fig:baseline} shows the comparison between the proposed tracker and baseline trackers on OTB-50 \cite{wu2013online} benchmark in terms of precision and success rate.
	According to the evaluation results shown in Figure \ref{fig:baseline}, we have the following observations: 1) By comparing Ours$_{hu}$ with KCF, the proposed tracker with the proposed Huber regularization term improves $3.1\%$ and $2.4\%$ in AUC score in terms of DP and OP, respectively, on the OTB-50 benchmark.
	2) By comparing Ours$_{hus}$, fDSST and KCF, the proposed tracker with scale variation outperforms $7.4\%$ and $9.9\%$ than the KCF in AUC score in terms of DP and OP, respectively.
	On the other hand, the fDSST only outperforms the KCF $3.4\%$ in terms of OP, while the precision is worse than KCF.
	3) A proper trade off between accuracy and speed is made. The computation cost of the proposed method is only increased a little as shown in the legend of Figure \ref{fig:baseline}, 197fps for Ours$_{hu}$ vs 252fps for KCF and 41.2fps for Ours$_{hus}$ vs 71.2fps for fDSST.
	
	\subsubsection{Quantitative Evaluation}
\begin{figure}[!htb]
		\centering
		\includegraphics[width=1.45\linewidth]{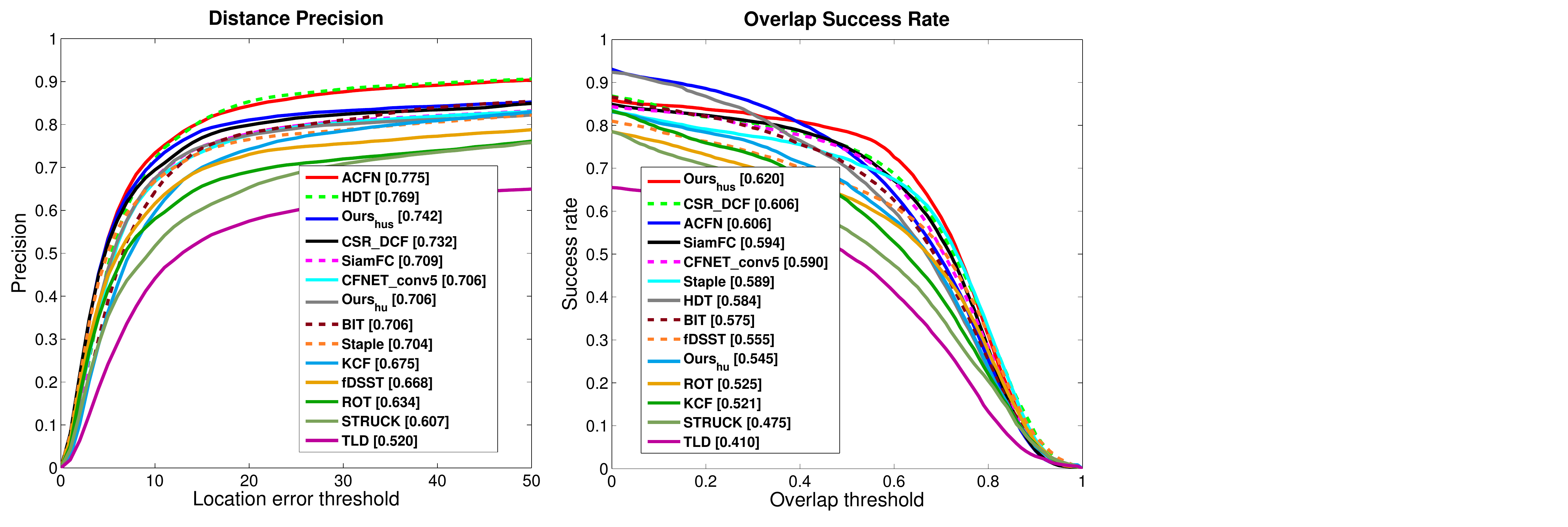}
		\caption{Quantitative results on the benchmark datasets. The scores in the legends indicate the average Area-Under-Curve values on all thresholds for success plots and precision plots, respectively.}
		\label{fig:overall}
	\end{figure}

	To validate overall performance for the proposed tracking approach, it is compared with 12 state-of-the-art trackers on OTB-50 \cite{wu2013online}, including five popular and diverse CF trackers (CSR\_DCF \cite{Lukežič2018}, Staple \cite{1Bertinetto2016}, ROT \cite{dong2017occlusion}, KCF \cite{1heri2015}, fDSST \cite{danelljan2017discriminative}, CSK \cite{1heri2013}), four deep learning based trackers (ACFN \cite{choi2017attentional}, SiamFC \cite{bertinetto2016fully}, CFNET \cite{valmadre2017end}, HDT \cite{qi2016hedged}) and four other trackers (BIT \cite{cai2016bit}, TLD \cite{1Kalal2012} and Struck \cite{1Hare2016}) which achieve favorable tracking performances on these two datasets.
	The precision and success plots over OTB-50 \cite{wu2013online} are presented in Figure \ref{fig:overall}.
	The detail information for each state-of-the-art trackers are listed in Table \ref{tab:fps}.
%	and Table \ref{tab:fps}
	According to Figure \ref{fig:overall}, it is observed that the proposed method achieves the best and third best AUC score in terms of OP and DP.
	Particularly, the ACFN (CVPR17) and HDT (CVPR16) outperforms $3.3\%$ and $2.7\%$ than Ours in terms of DP, while Ours outperforms $1.4\%$ and $3.6\%$ than ACFN and HDT in terms of OP, respectively.
	However, the speed of the proposed method is much faster than these two trackers, 41.2fps for Ours$_{hus}$ comparing with 15fps for ACFN and 10fps and for HDT, and is GPU independent while ACFN and HDT need the support of GPU. 
	On the other hand, for comparisons made under the metric of values at threshold of 20px and 0.5 for DP and OP respectively, as shown in Table \ref{tab:fps}, the proposed method achieves best for OP, third best for OP and best for the average of OP and DP, which demonstrates the effectiveness and efficiency of the solution for the proposed Huber regularization.
	\begin{table*}[h]
		\caption{A comparison of our approach using overlap precision (OP) with values at threshold of 0.5 and distance precision (DP) with values at threshold of 20px with the recent state-of-the-art trackers on OTB-50 \cite{wu2013online} dataset. The average speed (\ie, frames per second, FPS) is evaluated on the whole dataset. The first and second best scores are highlighted with \textcolor{red}{red} and \textcolor{blue}{blue}, respectively. The speed labeled with * represents the tracker reaches the requirement of real time tracking.}
		\begin{center}
			\renewcommand\arraystretch{1.2}
			\scalebox{0.84}{
				\begin{tabular}{cccccccccccccccc}
					\hline
					& ACFN & HDT & CSR\_DCF & SiamFC & CFNET & Staple & BIT & fDSST & ROT & KCF & STRUCK & TLD & CSK & Ours$_{hu}$ & Ours$_{hus}$\\
					\hline
					when & 2017 & 2016 & 2017 & 2016 & 2017 & 2016 & 2017 & 2017 & 2017 & 2015 & 2011 & 2011 & 2012 & & \\
					where & CVPR & CVPR & CVPR & ECCVW & CVPR & CVPR & TIP & TPAMI & TMM & TPAMI & ICCV & TPAMI & ECCV & & \\
					\hline
					Need GPU & Y & Y & N & Y & Y & N & N & N & N & N & N & N & N & N & N\\
					\hline
					OP (\%) & 74.8 & 70.9 & 75.4 & 75.4 & 74.6 & 72.5 & 71.6 & 66.5 & 63.7 & 63.3 & 56.2 & 50.8 & 44.5 & 67.0 &\textcolor{red}{78.8}\\
					DP (\%) & \textcolor{blue}{83.8} & \textcolor{red}{84.7} & 79.5 & 77.4 & 76.9 & 76.1 & 77.6 & 72.3 & 68.5 & 73.6 & 64.3 & 56.7 & 53.3 & 77.1 & \textbf{80.7}\\
					Mean (\%) & \textcolor{blue}{79.3} & 77.8 & 77.5 & 76.4 & 75.8 & 74.3 & 74.6 & 69.4 & 66.1 & 68.5 & 60.3 & 53.8 & 48.9 & {72.1} & \textcolor{red}{79.8}\\
					\hline
					Speed(FPS) & 15 & 10 & 6.8 & $ 86^* $ & $ 75^* $ & $ 67.7^* $ & $ 46.4^* $ & $ 71.2^* $ & $ 29.4^* $ & $ \textcolor{blue}{252}^* $ & 10.1 & 21.8 & $ \textcolor{red}{375}^* $ & $ 197^* $ &$ \textbf{41.2}^* $\\
					\hline
			\end{tabular}}
		\end{center}
		\label{tab:fps}
	\end{table*}

	\subsubsection{Qualitative Evaluation}
\begin{figure*}[!htb]
		\centering
		\includegraphics[width=0.99\linewidth]{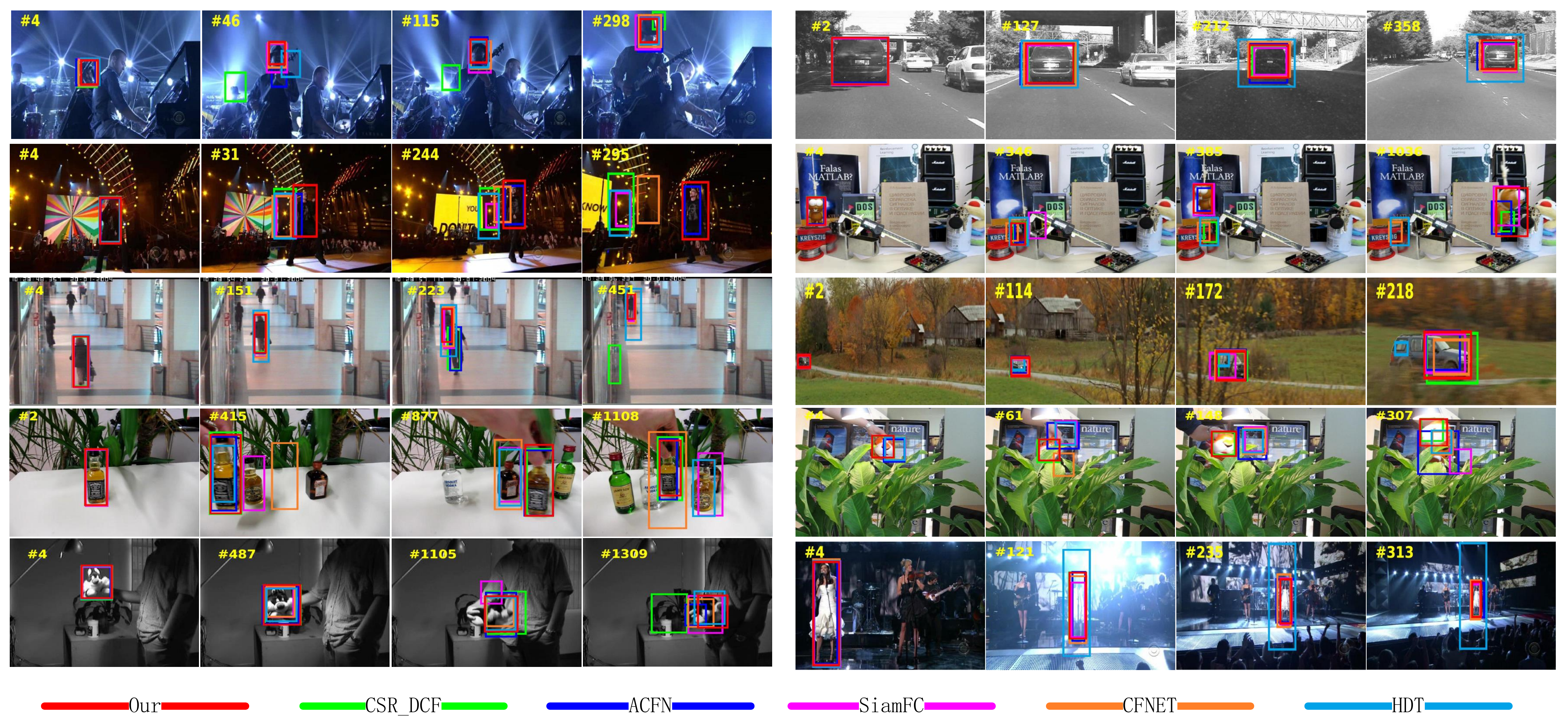}
		\caption{Bounding box comparison on ten challenging image sequences (from left to right and top to down are \textit{Shaking, Car4, Singer2, Lemming, Walking2, CarScale, Liquor, Tiger1, Sylvester, Singer1, respectively}).}
		\label{fig:qual}
	\end{figure*}

	Qualitative comparisons on a subset of sequences are also conducted.
	Though we compared with all the evaluated trackers, here we only present the bounding box results of top six trackers, namely CSR\_DCF \cite{Lukežič2018}, ACFN \cite{choi2017attentional}, SiamFC \cite{bertinetto2016fully}, CFNET \cite{valmadre2017end} and HDT \cite{qi2016hedged} with some selected frames in Figure \ref{fig:qual}. 
	As observed, the results demonstrate that the proposed tracker is able to locate the targets more precisely despite the existing of various challenges.
	For example, the sequences such as\textit{Lemming, Tiger1 and Walking2}, include some challenge attributes, such as occlusion, scale variation, \etal 
	The proposed tracker performs well in these sequences as shown in several representative frames while other trackers such as CFNET (CVPR17) drift when occlusion happens.
	Some sequences including illumination variation, such as \textit{Shaking, Singer2}, which will introduce unpleasant noise information and reduce the distinction between target and its surrounding backgrounds will obviously increase the difficulty of the tracking task.
	Fortunately, the proposed method can handle this challenge properly and achieves more accurate tracking result compared to other state-of-the-art trackers. 
	Tracking results on some other sequences, such as \textit{Sylvester, Car4, Singer1,} \etal also demonstrate that the proposed tracker can achieve higher accuracy than others.
	Actually, almost all the other trackers are unable to handle these complicated scenarios.
	\begin{figure*}[!htb]
		\centering
		\includegraphics[width=1.4\linewidth]{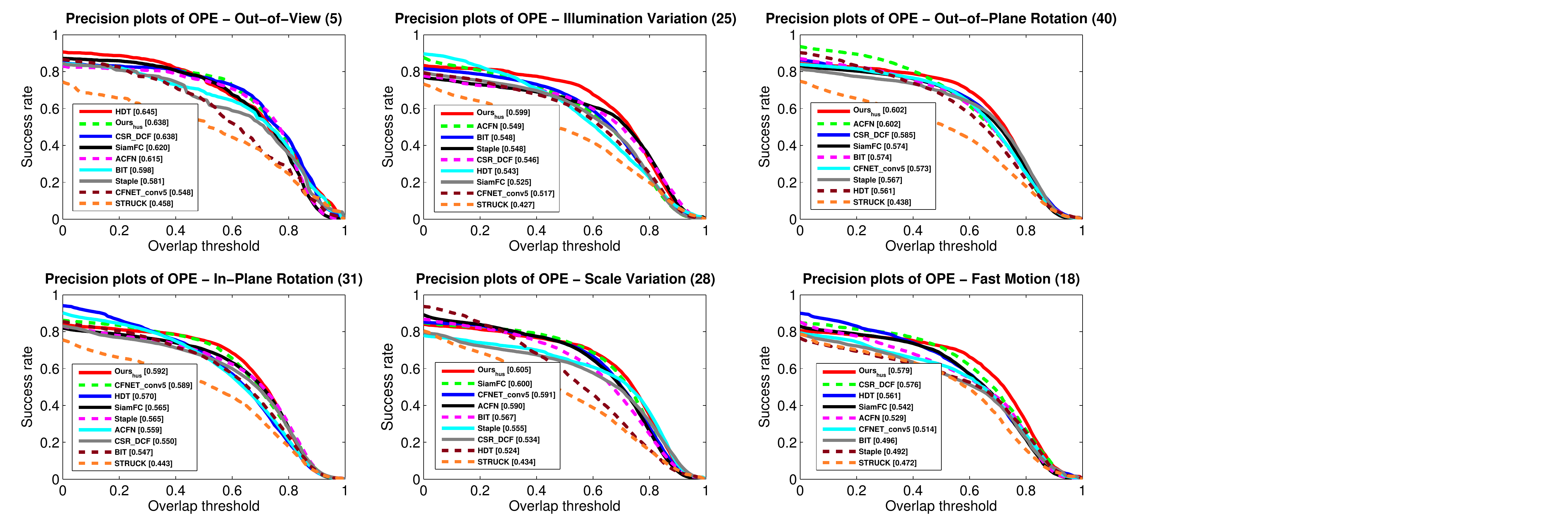}
		\caption{Quantitative comparison results on OTB-50 \cite{wu2013online} for six challenge attributes (from right to left and top to down are out-of-view, illumination variation, out-of-plane rotation, in-plane rotation, scale variation and fast motion, respectively). The value behind each attribute denotes the number of sequence that this attribute includes.
		}
		\label{fig:att}
	\end{figure*}

	\subsubsection{Comparisons on different attributes on OTB-50 \cite{wu2013online}}
	To thoroughly evaluate the robustness of the proposed visual tracker in various scenes, we present tracking accuracy in terms of 6 challenging attributes on OTB-50 \cite{wu2013online} in Figure \ref{fig:att}.
	As observed, the proposed tracker outperforms other methods by a huge margin in most of the attributes, particularly in handling illumination variation, which outperforms the second best tracker $5.0\%$.
	Note that this attribute introduces great challenge for the trackers due to the huge variation of target appearance and its surrounding backgrounds.
	It is important for a tracker to eliminate the noise information produced by lighting variation and correctly track the target if it could concentrate more on the target itself other than the noise background.
	
	Another case is deformation.
	That is, the target deforms locally in the appearance.
	This enlarges the representation error of the target since there is some background merged with the truth target in the preferred bounding box, enhancing the difficulty of tracking task.
	However, by introducing the proposed Huber regularization, the proposed tracker could diminish the distraction from target appearance variation and its surrounding background changing, due to the feature selection attribute of the sparsity constraint regularization term.
	
	\subsubsection{Comparisons on tracking speed}
	To demonstrate the efficiency of the proposed solution in the Fourier Domain for Huber regularization term, we implement the proposed tracker in two different versions as mentioned in Section \ref{com:baseline}.
	The results for all these trackers are listed in Table \ref{tab:fps}, based on which the following observations are made:
	a) The proposed tracker achieves the highest tracking accuracy, while it gives competitive tracking speed compared with other CF-based trackers such as CSR\_DCF (CVPR17), Staple (CVPR16), KCF (TPAMI15), fDSST (TPAMI17).
	b) The proposed method improves the tracking accuracy with little sacrifice on tracking speed when comparing Ours$_{hu}$ with the baseline tracker KCF.
	More specifically, the tracking accuracy is improved from $67.5\%$ to $70.6\%$ in terms of DP and from $52.1\%$ to $54.5\%$ in terms of OP, while the tracking speed is only decreased about $55$fps.
	c) The tracking speed of the proposed method Ours$_{hus}$ is about $3$ times faster than the second best tracker ACFN (CVPR17) and $4$ times faster than the third best tracker HDT (CVPR16) as shown in Table \ref{tab:fps}.

	\subsection{Evaluation on VOT16 \cite{Kristan2016a} Dataset}
	In addition to OTB-50 \cite{wu2013online}, we also evaluate the proposed method on VOT16 \cite{Kristan2016a}, which contains 60 challenging real-life videos with various challenges.
	We compare our tracker with the following state-of-the-art methods: DNT \cite{chi2017dual}, CTF \cite{rapuru2017correlation}, Staple\_CA \cite{mueller2017context}, LoFT\_Lite \cite{poostchi2016semantic}, SiamAN \cite{bertinetto2016fully}, DFST \cite{BMVC2016_120}, SCT \cite{choi2016visual}, SWCF \cite{gundogdu2016spatial}, sKCF \cite{montero2015scalable}, CMT \cite{nebehay2015clustering} and BDF \cite{maresca2014clustering}.
	The version of the proposed tracker used in this comparison is Ours$_{hus}$ which includes both the optimization for the proposed Huber regularization term and scale variation detecting.
	\begin{figure}[!htb]
		\centering
		\mbox{
			\subfigure{
				\includegraphics[width=0.85\linewidth]{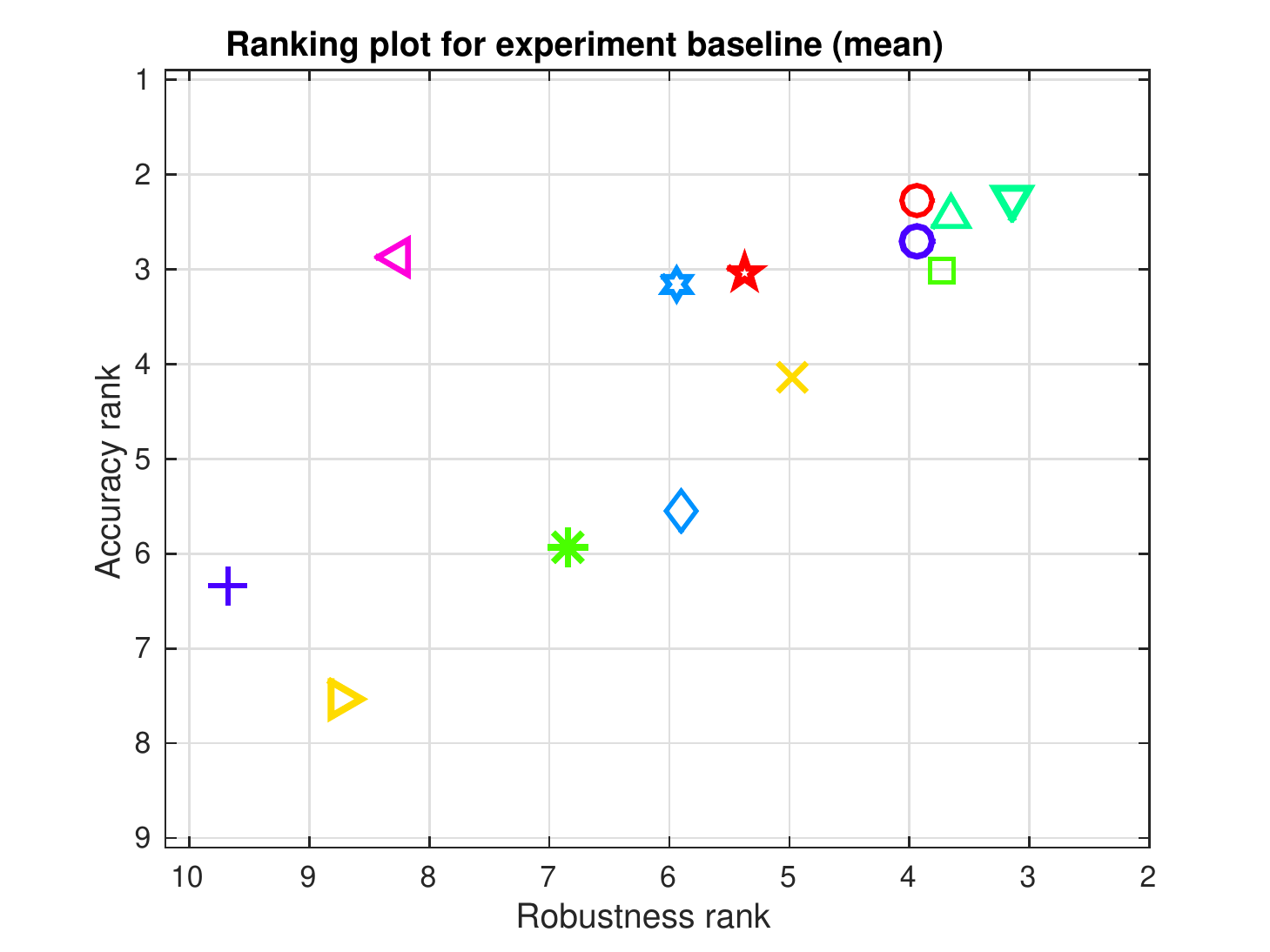}
		}}
		\mbox{
			\subfigure{
				\includegraphics[width=1\linewidth]{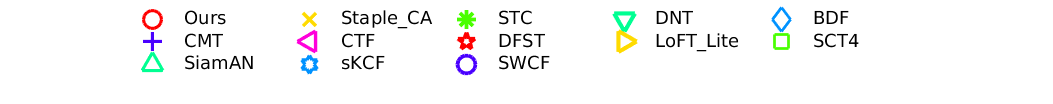}
			}
		}
		\caption{Accuracy-robustness ranking plot for the state-of-the-art comparison on VOT2016 \cite{Kristan2016a} dataset.}
		\label{fig:vot}
	\end{figure}

	\begin{table*}[h]
		\caption{Comparison with the state-of-the-art trackers on the VOT2016 dataset. The results are presented in terms of accuracy rank (Ar), robustness rank (Rr) and average ranking (Avg). \textcolor{red}{Red}: best. \textcolor{blue}{Blue}: second best.}
		\begin{center}
			\renewcommand\arraystretch{1.3}
			\scalebox{0.96}{
				\begin{tabular}{cccccccccccccc}
					\hline
					& DNT & SWCF & SCT & LoFT\_Lite & SiamAN & CTF & Staple\_CA & sKCF & DFST & STC & CMT & BDF & Ours$_{hus}$\\
					\hline
					when & 2017 & 2016 & 2016 & 2016 & 2016 & 2017 & 2017 & 2015 & 2016 & 2014 & 2015 & 2014 &  \\
					where & TIP & ICIP & CVPR & CVPRW & ECCVW & TIP & CVPR & ICCVW & BMVC & ECCV & CVPR & ECCVW &  \\
					\hline
					Need GPU & Y & N & Y & N & Y & N & N & N & N & N & N & N & N\\
					\hline
					Ar  & \textcolor{red}{2.25} & 2.80 & 3.08 & 7.53 & 2.62 & 2.88 & 4.13 & 3.17 & 3.05 & 5.93 & 6.33 & 5.55 & \textcolor{blue}{2.28}\\
					Rr  & \textcolor{red}{3.15} & 3.93 & 3.72 & 8.73 & \textcolor{blue}{3.65} & 8.27 & 4.97 & 5.93 & 5.37 & 6.85 & 9.68 & 5.90 & \textbf{3.94}\\
					\hline
					Avg  & \textcolor{red}{2.70} & 3.32 & 3.40 & 8.13 & 3.14 & 5.58 & 4.56 & 4.56 & 4.21 & 6.39 & 8.01 & 5.73 & \textcolor{blue}{3.11}\\
					\hline
			\end{tabular}}
		\end{center}
		\label{tab:vot2016}
	\end{table*}
	Among these competitive trackers, some of them are correlation filter based, such as Staple\_CA (CVPR17), CTF (TIP17), SWCF (ICIP16), and deep learning based, such as DNT (TIP17), SCT (CVPR16), SiamAN (ECCVW16).
	As shown in Figure \ref{fig:vot} and Table \ref{tab:vot2016}, the proposed tracker achieves the second best in terms of accuracy, the fifth in terms of robustness and the second in average rank.
	Compared with the best tracker DNT, the proposed tracker is closed to it in both the accuracy and robustness. 
	Such small deviations are acceptable in many real time applications, such as robotic navigation without a precise map. 
	On the other hand, DNT is a GPU dependent tracker and its tracking speed is only about 5 frames per second, while the proposed one runs as efficiently as 41.2 frames per second without the support of GPU.
	Overall, the proposed tracker achieves a good trade-off among tracking speed, robustness and accuracy, which makes it competitive to other state-of-the-art trackers.
	
	\subsection{Limitation of the Proposed Tracker}
	As shown in the success plots of out-of-view in Figure \ref{fig:att}, the proposed method only achieves the second best rank,	since the proposed tracker focuses on improving the tracking performance through introducing the feature selective ability instead of handling the case of recovering the tracker from target disappearance when the target moves out of the camera's view.
	Nevertheless, the proposed tracker still achieves better accuracy compared with other correlation filter based trackers, such as CSR\_DCF (CVPR17), Staple (CVPR16).
	However, the proposed tracker may not perform well for some extremely challenging cases such as the $rabbit$ in VOT16 \cite{Kristan2016a} and the $Matrix$ in OTB-50 \cite{wu2013online}.
	For these tracking failure sequences, the target is quite similar to interference of surroundings, resulting in that the proposed tracker cannot learn a discriminate model to prominent the target while weakening the background part.
	Especially for the sequence \textit{Matrix}, it also contains fast motion, which is a huge challenge attribute for correlation filter based trackers due to the limited regression region.
	In fact, the aforementioned sequences are so challenging that most of the trackers cannot perform well.
	Nevertheless, the comparison results on these two large datasets demonstrate that the proposed tracking approach can achieve robustly tracking performance among various sequences with different targets.
	As such, higher tracking accuracy is achieved while the tracking speed is only reduced slightly.
	In conclusion, based on the experimental results, the proposed method is proven to be sufficiently effective and efficient to handle various environmental challenges.

	\section{Conclusions}
	\label{sec:conclusion}
	In this paper, we have presented a novel and fast solution for spatial constraint optimization problem by formulating the original problem in the Fourier Domain, and achieved a simple closed-form solution for this optimization problem with the attribute of Convolution Theorem.
	Our tracking algorithm outperforms the state-of-the-art trackers in terms of both tracking speed and accuracy, in OTB-50 \cite{wu2013online} and VOT16 \cite{Kristan2016a} benchmarks.
	
	The proposed tracker is very useful in practical applications, for example autonomous navigation without a precise map. Visual inertial odometry (VIO) will be adopted to roughly determine the location of an agent \cite{forster2017manifold}.
	Visual place recognition (VPR) can be utilized to remove the drift on the VIO \cite{lowry2016visual}.
	The proposed tracker can be adopted to determine the moving direction for the agent.   The final accuracy can be preserved using visual servoing with an advanced controller \cite{Li2018Asymptotic}.
	Such issues will be addressed in our future research.
	\section*{Acknowledgement}
	
	This research is supported by SERC grant No. 162 25 00036 from the National Robotics Programme (NRP), Singapore.
	
%	\bibliographystyle{IEEEtranTIE}
%	\bibliography{IEEEabrv,BIB_FDKCF}\ %IEEEabrv instead of
	
\end{document}